\documentclass{article}

\usepackage[preprint]{neurips_2025}

\usepackage[utf8]{inputenc} 
\usepackage[T1]{fontenc}    
\usepackage{hyperref}       
\usepackage{url}            
\usepackage{booktabs}       
\usepackage{amsfonts}       
\usepackage{nicefrac}       
\usepackage{microtype}      
\usepackage{xcolor}         

\usepackage{graphicx} 
\usepackage{subcaption}

\usepackage{enumitem}

\title{Associative Syntax and Maximal Repetitions reveal context-dependent complexity in fruit bat communication}

\author{%
  Luigi Assom \\
  Alumnus, Department of Computer and Systems Sciences, Stockholm University \\
  Stockholm, Sweden \\
  \texttt{luigi.assom@gmail.com} \\
}

\begin{document}

\maketitle

\begingroup
\renewcommand\thefootnote{}\footnote{
This work was presented at the 39th Conference on Neural Information Processing Systems (NeurIPS 2025) Workshop on AI for Non-Human Animal Communication (non-archival).
}
\addtocounter{footnote}{-1}
\endgroup

\begin{abstract}
   This study presents an unsupervised  method to infer discreteness, syntax and temporal structures of fruit-bats vocalizations, as a case study of graded vocal systems, and evaluates the complexity of communication patterns in relation with behavioral context. The method improved the baseline for unsupervised labeling of vocal units (i.e. syllables) through manifold learning, by investigating how dimensionality reduction on mel-spectrograms affects labeling, and comparing it with unsupervised labels based on acoustic similarity. We then encoded vocalizations as syllabic sequences to analyze the type of syntax, and extracted the Maximal Repetitions (MRs) to evaluate syntactical structures. We found evidence for: i) associative syntax, rather than combinatorial (context classification is unaffected by  permutation of sequences, $F1 > 0.9$); ii) context-dependent use of syllables (Wilcoxon rank-sum tests, \textit{p-value < 0.05}); iii) heavy-tail distribution of MRs (truncated power-law, exponent $\alpha < 2$), indicative of mechanism encoding combinatorial complexity. Analysis of MRs and syllabic transition networks revealed that mother-pupil interactions were characterized by repetitions, while communication in conflict-contexts exhibited higher complexity (longer MRs and more interconnected vocal sequences) than non-agonistic contexts. We propose that communicative complexity is higher in scenarios of disagreement, reflecting lower compressibility of information.
  
\end{abstract}

\section{Introduction}

Quantifying communication complexity in species with graded vocal systems remains a key challenge. We improved and extended an unsupervised pipeline to infer repertoire and syntax from vocalizations, applying it to fruit bats as a case study. We propose Maximal Repeats (MRs) as a novel metric to capture combinatorial complexity, extending variables of communication complexity rooted in information-theory to avoid non-independency between communication and sociality, which is a circularity pitfall in the social complexity hypothesis for communication complexity (SCHCC)  \cite{peckre2019clarifying}.

Current methods face limitations. Sainburg et al. \cite{sainburg2020finding} \cite{sainburg2021toward} use manifold learning \cite{sainburg2021parametric} to cluster vocal units, but this approach assumes discrete systems with clear unit boundaries and struggles with the continuous, graded vocalizations of species like fruit bats. Zhang et al. \cite{zhang2019comparing} analyze syntax of horseshoe bats using behavioral classifiers discriminating between aggressive and distressing calls, but require ground-truth syllable labels from experts, limiting scalability to other behavioral contexts and to other species.

Our work addresses two research questions:
\begin{itemize}
    \item RQ1: How does dimensionality reduction affect unsupervised clustering on manifold learning for quantifying size and diversity of the repertoire ?
    \item RQ2: How do syntax and temporal structure encode contextual information?
\end{itemize}

To answer these, we first refined the method of Sainburg et al. \cite{sainburg2020finding} by inspecting how dimensionality reduction of mel-spectrograms affects clustering in manifold learning for the unsupervised labeling task; then, we used the labels to encode vocalizations as sequences and engineer features of a behavioral classifier, based on Zhang et al. \cite{zhang2019comparing}, to test if order of syllables affect classification (i.e. compositional or associative type of syntax \cite{suzuki2020syntax}). We extended the work with sequence analysis and introduce MRs - to our knowledge novel to animal communication - as a variable to analyze combinatorial complexity, motivated by their application in computational linguistics \cite{dkebowski2015maximal} and inspired by the analogical problem of how limited repertoires encode complex information in genetics (e.g. nucleotides in DNA sequences for protein expressions).

We used the fruit bat dataset \cite{prat2017annotated} as a case study because it can be compared with the clustering baseline in \cite{sainburg2020finding} and because their authors provide domain reference of repertoire size and syntax-type for evaluating our unsupervised results \cite{amit2023bat}.

Our contributions are:
\begin{enumerate}
    \item A refined unsupervised pipeline for repertoire quantification in graded vocal systems, improving upon \cite{sainburg2020finding} and yielding results consistent with expert knowledge \cite{amit2023bat}.
    \item An analysis of context-dependent syntax, adapting the method of \cite{zhang2019comparing} to use syllables automatically labeled in multiple context-dependent repertoires.
    \item The novel application of Maximal Repeats to animal communication, providing evidence for heavy-tailed distributions and proposing MR length as a metric of complexity.
    \item Findings suggesting higher communicative complexity (longer MRs) in conflict behaviors versus cooperative ones.
\end{enumerate}

We anticipate the limitation that the terms "conflictual" or "cooperative" are our interpretations of the behavioral annotation in the original dataset \cite{prat2017annotated}.

\section{Background and Motivation}

\textbf{Discreteness in graded vocal systems }
A key challenge in deciphering animal communication is identifying the linguistic units relevant to the species, addressing discreteness as basis for syntax \cite{andreas2022toward}. A method for unsupervised labeling clusters spectrogram representations of acoustic segments through manifold learning \cite{sainburg2020finding} \cite{sainburg2021toward}, assuming that: i) vocal units are characterized by independent time-frequency features (e.g. clear unit boundaries as in discrete vocal systems); ii) acoustic similarity between units is relevant to the species \cite{sainburg2020finding}. However, in graded vocal systems the time-frequency features overlaps between syllables, and thus clustering performance degrade \cite{peckre2019clarifying}. Poor efficacy on fruit bat vocalizations and other graded systems (e.g. mice, human phonemes) \cite{sainburg2020finding}, prompted our study on how dimensionality reduction of input spectrograms may mitigate the challenges posed by vocal gradation. 

\textbf{Communication complexity \& social complexity.}
The social complexity hypothesis for communicative complexity (SCHCC) recommends using quantitative information-theoretic metrics, such as the number of signaling units and the variety of their assembling patterns, to gauge communication complexity and to mitigate risks of non-independence between sociality and communication variables \cite{peckre2019clarifying}. However, even if defining units through information theory helps to align definitions across fields (e.g. biology; linguistics), quantifying information does not necessarily imply meaning \cite{kershenbaum2016acoustic}.

\textbf{Syntax as systems conveying meaning.}
Syntax and temporal organization are considered systems that convey meaning \cite{berthet2023animal}. The combination of units can form associative (order-independent) or combinatorial (order-dependent) syntax, which can yield to recursion or idiomatic types \cite{suzuki2020syntax}. Heavy-tail distributions of signals (e.g. Zipf's law) are clues of compression mechanisms that minimize communication costs and imply rudimentary syntax \cite{i2005consequences}. Shannon entropy is a robust estimator of Zipf’s power-law coefficient to assess repertoire variability (i.e. potential information content) \cite{kershenbaum2021shannon}, measuring uncertainty in repertoire draws \cite{peckre2019clarifying}. However, it doesn't fully capture long-range dependencies and combinatorial capacity of a system.

\textbf{Maximal Repeats: A metric for combinatorial complexity.}
These dependencies can be explored in terms of information-decay. Comparative studies show that in birdsong and human speech information decay is exponential for short sequences (e.g. notes) but follows power laws for longer ones (e.g. syllables or phrases); in humans, this mechanism predates the acquisition of language \cite{sainburg2019parallels}, implying potential common aspects across species phylogenetically distant. Maximal repetitions (MRs) offers a complementary tool. In computational linguistics, 
the scaling of MRs is mathematically linked to block entropy (entropy of \(n\)-grams); in written texts, the MR length was found to grow as a power of the logarithm of text length, consistent with Hilberg’s conjecture, which posits that block entropy grows sub-linearly as a power-law and supports the view that natural language possesses strong long-range dependencies and is highly compressible \cite{dkebowski2015maximal}. We did not find studies employing MRs in animal communication and propose its use for estimating the combinatorial capacity of a system, thus extending variables of communication complexity in \cite{peckre2019clarifying}.

\section{Methodology}
We designed two experiments: unsupervised labeling to infer repertoire size and diversity (addressing RQ1); behavioral classification and statistical analysis of syllabic sequences to infer syntax type and temporal structures across behaviors (addressing RQ2).
 
\textbf{Dataset.} We used the annotated fruit bat vocalization dataset from \cite{prat2017annotated}, featuring 41 specimens with emitter, addressee, and behavioral context labels. Vocal units were automatically segmented by the authors. We analyzed vocalizations from the contexts: Mating Protests; Fighting; Threat-like; Biting; Feeding; Grooming; Kissing; Isolation (meaning mother-pup interactions). Contexts labeled as: Generic, Sleeping (utterances in sleeping area), or Unknown, were excluded due to ambiguity.

\subsection{Size and Diversity of Repertoire}
This experiment evaluated how the dimensionality of mel-spectrograms affects unsupervised clustering performance for quantifying the repertoire. The pipeline follows \cite{sainburg2020finding}:  spectrograms (or their representations from autoencoders (AEs) \cite{sainburg2021parametric}) are projected into a low-dimensional space with Uniform Manifold Approximation and Projection (UMAP), and then clustered with Hierarchical Density-Based Spatial Clustering (HDBSCAN).

\begin{figure}[htbp]
    \centering
    \begin{subfigure}[b]{\textwidth}
        \centering
        \includegraphics[width=0.6\linewidth]{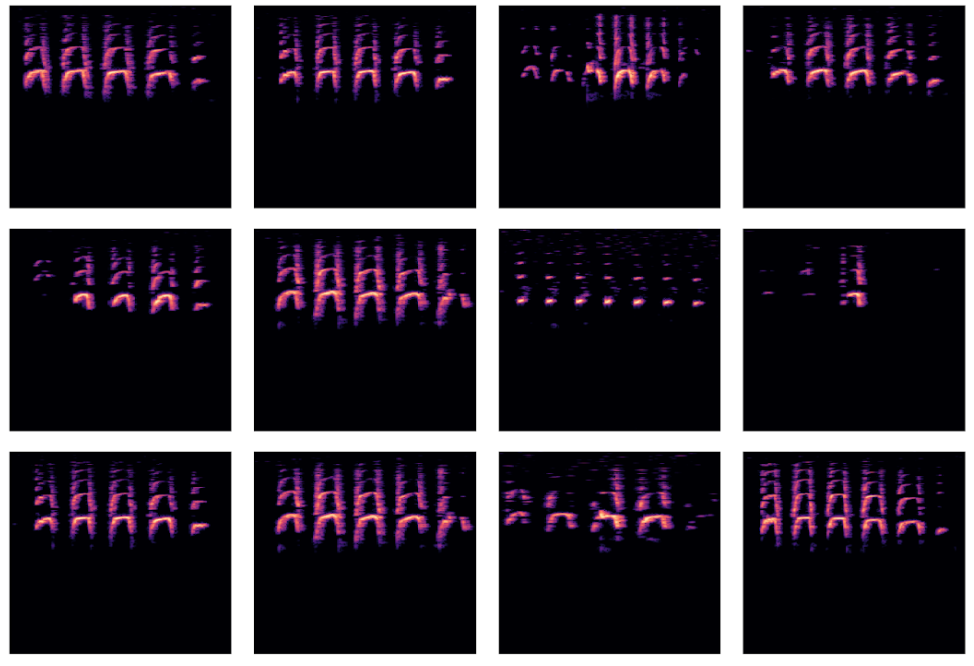}
        \caption{Individual syllable instance.}
        \label{fig:isolation_syllable_single}
    \end{subfigure}
    
    \vspace{0.5cm} 
    
    \begin{subfigure}[b]{\textwidth}
        \centering
        \includegraphics[width=0.9\linewidth]{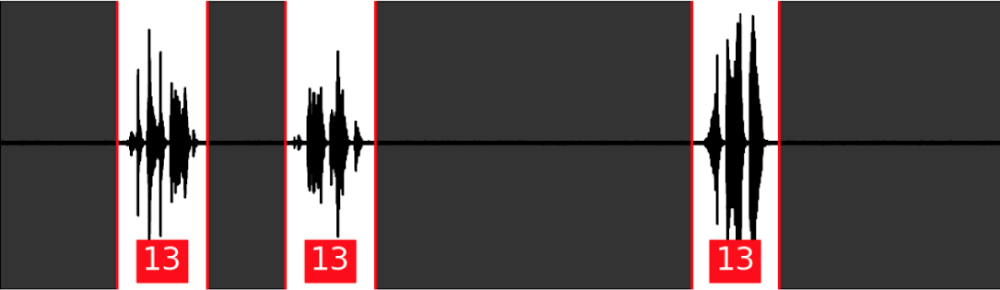}
        \caption{Sequence of repetitive occurrences.}
        \label{fig:isolation_syllable_sequence}
    \end{subfigure}
    
    \caption{Syllable-type unique to the Isolation context (mother-pupil interactions), isolated through agglomerative clustering. (\subref{fig:isolation_syllable_single}) Randomized sampling of the syllable-type,  displaying its uniform spectral structure. (\subref{fig:isolation_syllable_sequence}) A sequence of this syllable, demonstrating the characteristic repetition patterns consistent with underdeveloped vocalizations.
    \emph{Note --} Unsupervised labeling in these figures used the acoustic segments from the original dataset \cite{prat2017annotated}, whose boundaries were computed by thresholding the amplitude envelope above a fixed noise floor. When using algorithms that estimate the noise floor dynamically (as in \cite{sainburg2019parallels}), syllables could be further subdivided into smaller segments; in this example, the three bursts visible in the waveforms were separated into distinct sub-units. }
    \label{fig:isolation_syllable_composite}
\end{figure}

We systematically varied the dimensionality of input representations to explore the clustering performance on graded vocalizations, through: i) Spectrogram settings (probing extreme time–frequency trade-offs to test if separability of clusters stems more from time or frequency); ii) Dimensionality Reduction (using PCA on AEs latent representations of spectrograms, like in \cite{amit2023bat}, and testing different AEs architectures); iii) Segmentation (comparing the original procedure, which segmented audio where the amplitude envelope is above a fixed noise floor  \cite{prat2017annotated}, with Dynamic Threshold Segmentation, which estimates the noise floor dynamically and is helpful to isolate shorter sub-units \cite{sainburg2019parallels}) (see: settings for audio pre-processing in Table~\ref{tab:audio_preprocessing_params} and comments in Fig~\ref{fig:isolation_syllable_sequence}). This analysis was conducted on the top-5 emitters; the best configuration was scaled to the full dataset.

\textbf{Evaluation.} We used a two-tiered strategy due to the lack of ground-truth labels:
\begin{enumerate}[noitemsep,topsep=0pt]
    \item \textit{Internal Validation:} Silhouette Score to measure HDBSCAN cluster consistency.
    \item \textit{Agreement with Acoustic Similarity:} We generated a proxy for ground truth: for each emitter, we computed a pairwise distance matrix using Dynamic Time Warping (DTW) on Mel-Frequency Cepstral Coefficients (MFCCs) and performed Agglomerative Clustering with a quantile distance threshold (\(q=0.05\)). This yielded $27 \pm 2$ syllable types per emitter, consistent with known bat repertoire sizes \cite{amit2023bat}\cite{zhang2019comparing}. We measured agreement between these acoustic labels and HDBSCAN labels using the Adjusted Rand Index (ARI) and Normalized Mutual Information (NMI).
\end{enumerate}

\subsection{Type of syntax and temporal structures conveying contextual information }
\begin{figure}[htbp]
    \centering
    \includegraphics[width=0.9\linewidth]{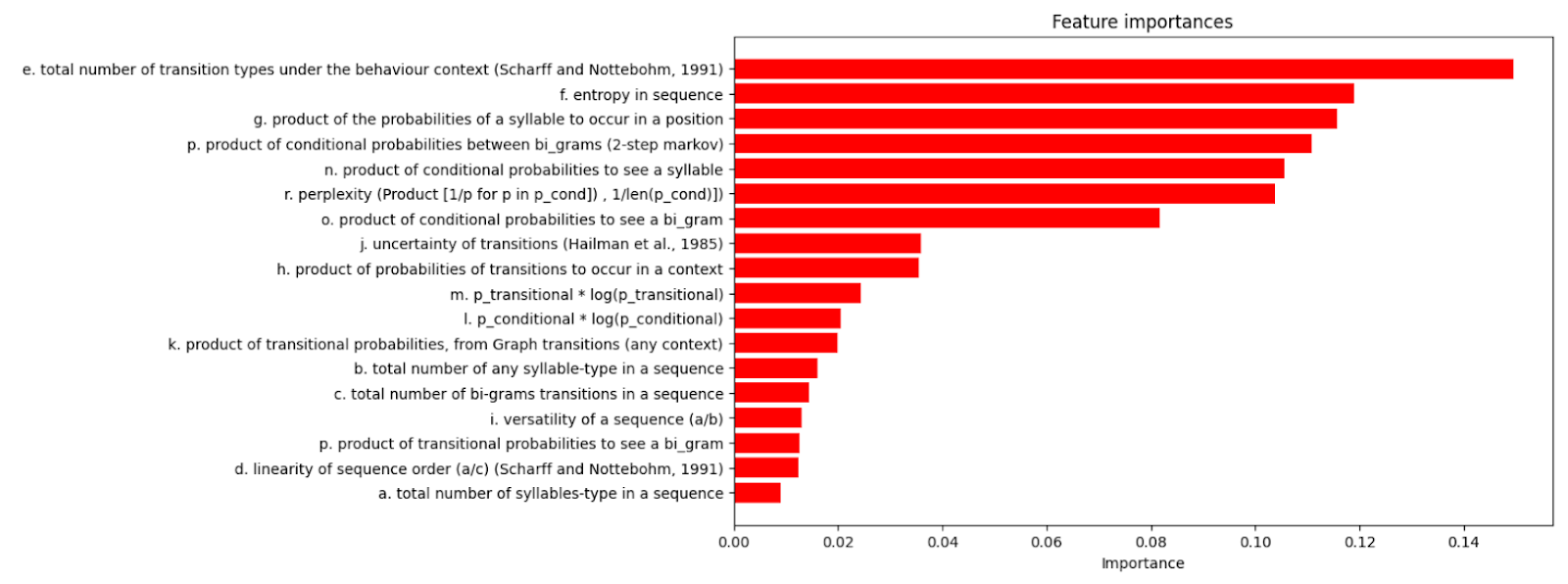}
    \caption{Importance of features used for the Random Forest classifier. Features representing richness of contextual syntax, unpredictability of sequences, commonness of patterns and strength of short transitions (respectively, features: \emph{e}, \emph{f}, \emph{g}, \emph{p}) account for about 50\% of the total feature importance, suggesting a predominant temporal organization of short transitions and repetitive patterns.}
    \label{fig:rf_feature_importance}
\end{figure}

This experiment investigated: 1) syntax type (associative/combinatorial), 2) context-dependent syllable usage, and 3) the distribution of Maximal Repeats (MRs). We tested three null hypotheses:
\begin{description}
    \item[HP1\textsubscript{0}:] \textit{Syllable order does not affect context classification.} \\
    \textbf{Method:} We replicated the Random Forest (RF) in \cite{zhang2019comparing} to classify behavior based on features from syllabic sequences (see predictors and their importance in Table~\ref{tab:rf_features}, Fig~\ref{fig:rf_feature_importance}). Unlike the original work, we used syllables from our unsupervised labels and extended the analysis to multiple behavioral classes. \\
    \textbf{Evaluation:} Comparison of $F1-scores$ between permuted and original sequences.
    
    \item[HP2\textsubscript{0}:] \textit{Syllable usage is identical across behaviors.} \\
    \textbf{Evaluation:} Wilcoxon rank-sum test on the syllable frequency distributions between pairs of behaviors.
    
    \item[HP3\textsubscript{0}:] \textit{The distribution of maximal repetitions follows an exponential distribution.} \\
    \textbf{Method:} We extracted MRs - the longest repeating subsequences - using a prefix-suffix tree algorithm. An exponential distribution would signify simple memory-less information decay (lower probability to observe longer sequences). A heavy-tailed distribution (e.g. power-law) would signify long-range dependencies \cite{dkebowski2015maximal}. \\
    \textbf{Evaluation:} Likelihood ratio test (exponential vs. power-law).
\end{description}

Finally, we compared the mean MR length across behaviors and qualitatively inspected the syllabic transition networks.

\section{Results}

\begin{figure}[htbp]
    \centering
    \begin{subfigure}[b]{0.48\textwidth}
        \centering
        \includegraphics[width=\textwidth]{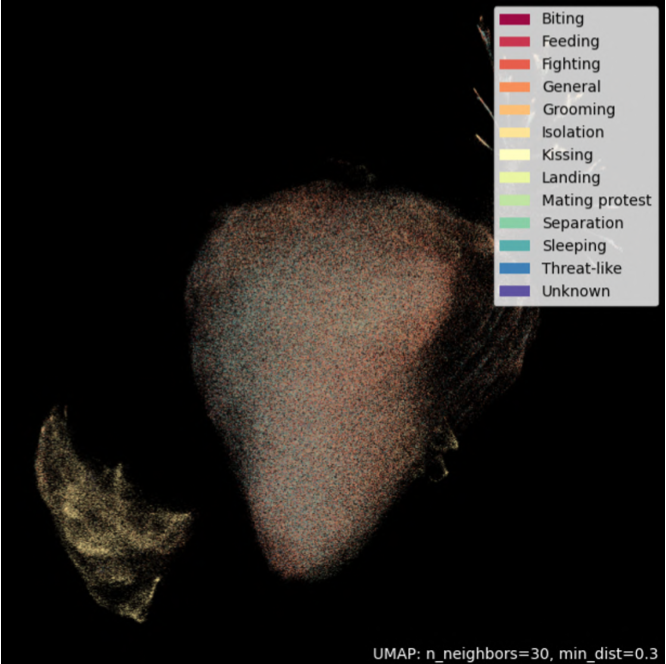}
        \caption{Benchmark results (replicated from \cite{sainburg2020finding}).}
        \label{fig:original_benchmark}
    \end{subfigure}
    \hfill
    \begin{subfigure}[b]{0.48\textwidth}
        \centering
        \includegraphics[width=\textwidth]{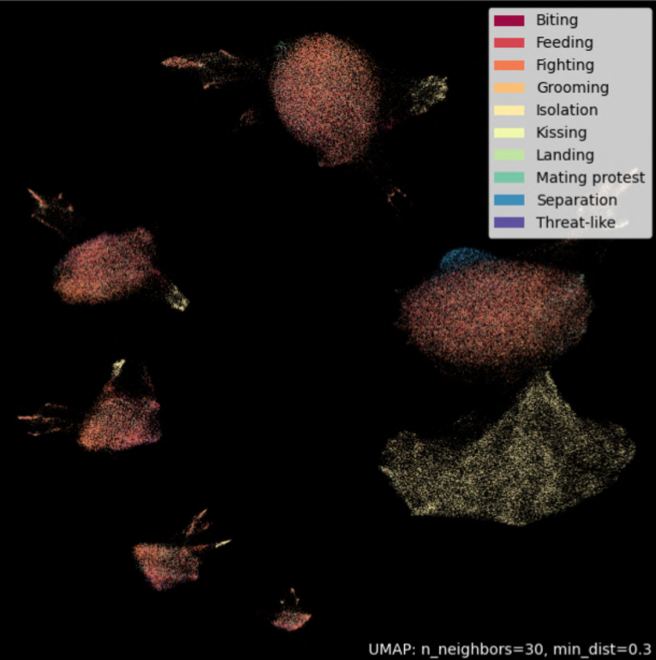}
        \caption{Our improved results with refined pipeline.}
        \label{fig:improved_benchmark}
    \end{subfigure}
    \caption{Improved clustering quality of continuous-type vocalizations. The left panel shows the original benchmark results, which primarily separate isolation calls from adult vocalizations. The right panel demonstrates our improved clustering, which identifies seven distinct syllable types through optimized dimensionality reduction and segmentation techniques applied to the graded vocal system.}
    \label{fig:clustering_improvement}
\end{figure}

\begin{figure}[htbp]
    \centering
    \begin{subfigure}[b]{0.48\textwidth}
        \centering
        \includegraphics[width=\textwidth]{images/improved_benchmark.png}
        \caption{} 
        \label{fig:improved_benchmark_side}
    \end{subfigure}
    \hfill
    \begin{subfigure}[b]{0.48\textwidth}
        \centering
        \includegraphics[width=\textwidth]{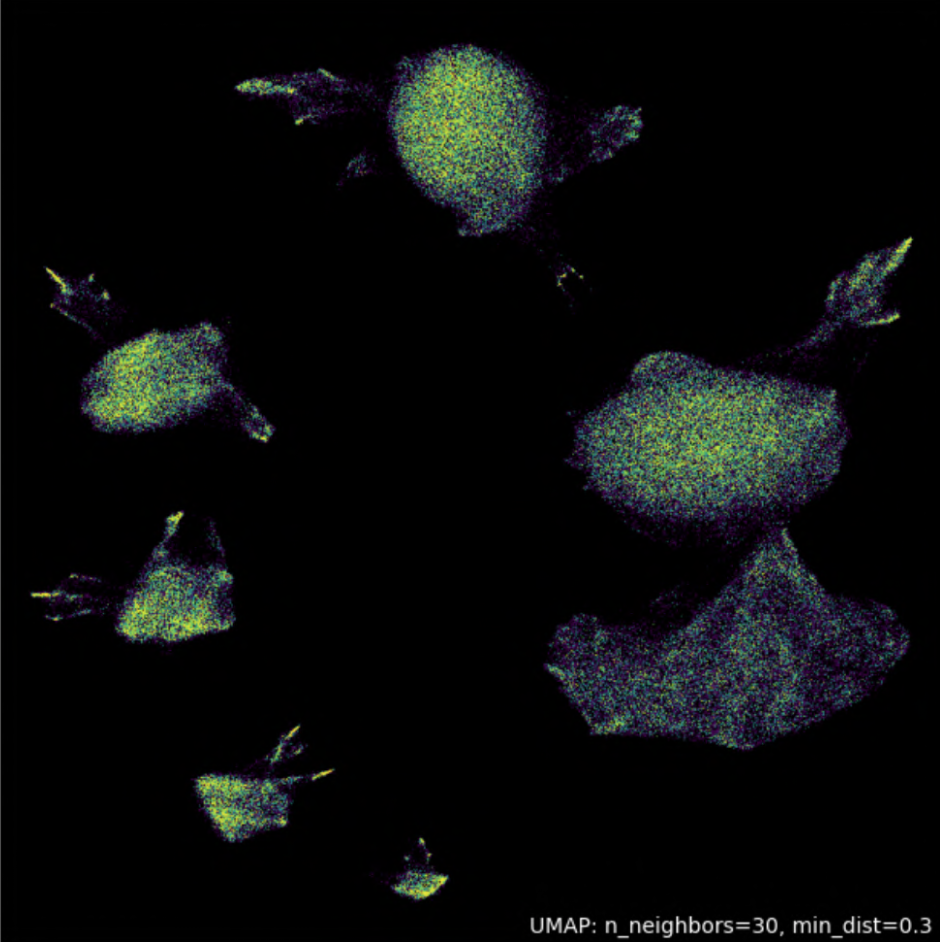}
        \caption{} 
        \label{fig:diagnostic_umap_side}
    \end{subfigure}
    \caption{Diagnostic of the local dimensionality of manifold learning. Clustering obtained from Non-parametric UMAP applied on Mel-Spectrograms (6x32) preprocessed by Mel-filterbank (hop size equal to FFT length) and dynamic segmentation. Inputs: 152,578 data-points from all bats (41 individuals). Bluish colors represent the lowest local dimensionality, which corresponds to underdeveloped vocalizations from the Isolation context (i.e., simpler, more uniform spectrograms). Warmer colors (yellow/red) indicate regions of higher local dimensionality and greater acoustic complexity.}
    \label{fig:diagnostic_composite}
\end{figure}

Cluster quality addressed the first question (RQ1). Coarse-graining the temporal dimension of spectrograms from vocal units segmented with dynamic segmentation yielded the best results ($Silhouette > 0.5$, 95\% assignment accuracy; see: Fig~\ref{fig:improved_benchmark}, Appendix), identifying seven types of vocal units and improving the previous baseline that discriminated only two (i.e.  utterances of mother and pups in Isolation and utterances between adults in all the other contexts).
The local dimensionality of the UMAP embedding, inspected with diagnostic tools, is visible in Fig~\ref{fig:diagnostic_composite} (Appendix) along with the spectrogram settings used.
The acoustic similarity proxy (Agglomerative Clustering on DTW distance) yielded an average of $27 \pm 2$ syllable types per emitter, consistent with known fruit bat repertoire sizes \cite{amit2023bat} \cite{zhang2019comparing}. The agreement between this proxy and our best HDBSCAN clustering, using mel-spectrograms retaining higher dimensionality as in the original experiment, was moderate (Mean ARI = $0.12 \pm 0.01$, Mean NMI = $0.30 \pm 0.01$), and suggested a repertoire of 14 syllables.

Results on syntax and temporal structure (RQ2) are as follows: 

\textit{Syntax Type (HP1).} The permutation test revealed that syllable order did not affect classification performance ($F1-score > 0.9$ for both original and permuted sequences). Failing to reject $HP1_0$ supports an \textit{associative} rather than combinatorial type of syntax, consistent with findings in \cite{amit2023bat}.

\textit{Syllabic distribution (HP2).}  Syllable distribution was significantly different between Isolation and other contexts ($p < 0.05$, Wilcoxon rank-sum test), aligning with observations in \cite{amit2023bat}. Although specific outcomes were dependent on the clustering methods defining the repertoire, we found no significant evidence to reject $HP2_0$ for the cooperative contexts of Feeding, Grooming, and Kissing in the majority of pairwise comparisons, suggesting more uniform syllable usage across these behaviors. Heatmaps of syllabic distribution also suggested that Emitters grew in the same colony may not have a different use of syllables.

\textit{Maximal Repeats Distribution (HP3).} The likelihood ratio test rejected $HP3_0$ ($p < 0.05$). The distribution of MR lengths was best described by a truncated power-law ($\alpha = 1.79$), indicating a heavy-tailed distribution inconsistent with a memory-less process and instead indicative of long-range temporal structures, reflecting combinatorial capacity of syntactical patterns.

\begin{figure}[htbp]
    \centering
    \includegraphics[width=0.9\linewidth]{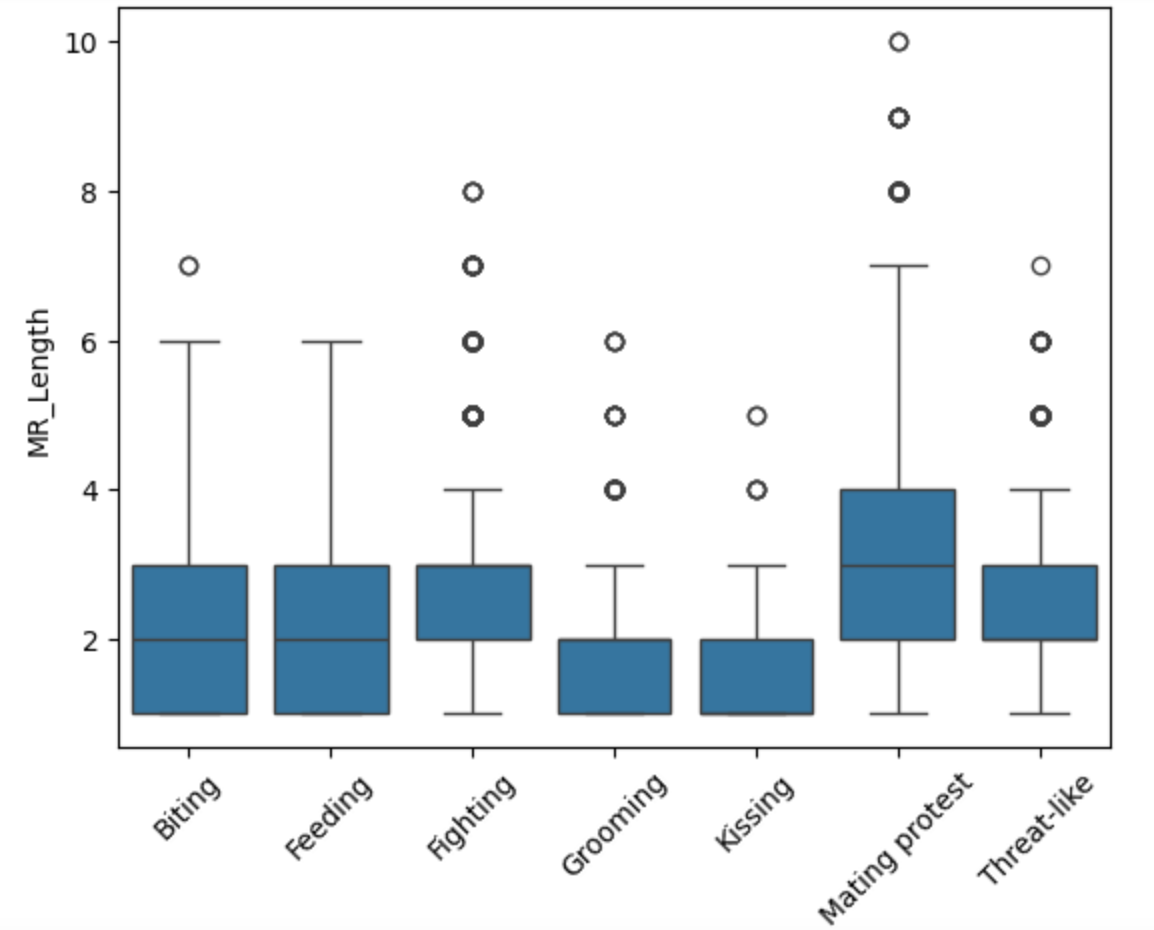}
    \caption{Distribution of Maximal Repetition (MR) lengths across behavioral contexts (sequences with at least 50 support). Conflict-related contexts (Mating Protest, Fighting, Threat-like) show heavier-tailed distributions with longer MRs, indicating more complex temporal structures and lower compressibility of information. Cooperative contexts (Feeding, Grooming, Kissing) exhibit shorter MR distributions, suggesting higher redundancy and more compressible communication patterns. The Isolation context shows a unique pattern dominated by short, repetitive sequences.}
    \label{fig:mr_distribution_behavior}
\end{figure}

\begin{figure}[htbp]
    \centering
    \begin{subfigure}[b]{0.45\textwidth}
        \centering
        \includegraphics[width=\textwidth]{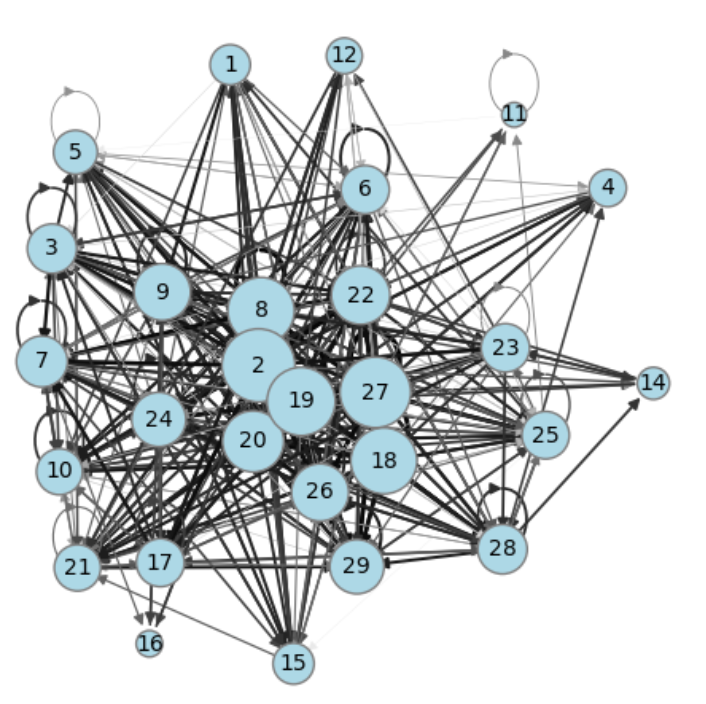}
        \caption{Mating Protest context.}
        \label{fig:nw_mating_protest}
    \end{subfigure}
    \hfill 
    \begin{subfigure}[b]{0.45\textwidth}
        \centering
        \includegraphics[width=\textwidth]{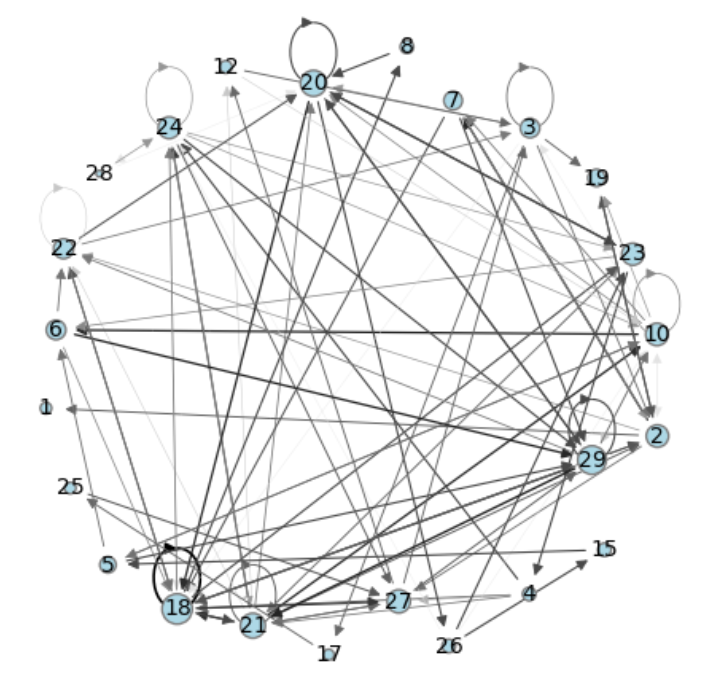}
        \caption{Kissing context.}
        \label{fig:nw_kissing}
    \end{subfigure}
    \caption{Examples of networks of syllabic transitions for Emitter ID 215 (syllables based on agglomerative clustering). These networks visually represent the transition probabilities between different syllable types within a specific behavioral context. The network structure for Mating Protest (\subref{fig:nw_mating_protest}) is denser and more interconnected, indicative of higher complexity, contrasting with the sparser structure of the Kissing context (\subref{fig:nw_kissing}).}
    \label{fig:network_examples}
\end{figure}

\textit{Behavioral Complexity through MRs and Networks.} The average length of MRs was greater in conflict-related contexts (Mating Protest, Fighting, Threat-like) than in cooperative ones (see: Fig~\ref{fig:mr_distribution_behavior}, Appendix). To further explore this complexity, we represented syllabic transitions as networks for each behavior. Quantitative analysis of these networks revealed a spectrum of structural properties: Conflict-related contexts exhibited network metrics indicative of a small-world architecture ($\omega \approx 0$), characterized by high local clustering (Avg $C > 0.4$) alongside efficient global connectivity; in contrast, cooperative contexts displayed metrics suggesting a more random, less structured network ($\omega > 0.5$) (see: Table~\ref{tab:graph_metrics_bat215}, Fig~\ref{fig:nw_mating_protest}, Appendix). Qualitatively,  graphs from the Isolation context showed simple repetitions of a specific syllable (see: Fig~\ref{fig:isolation_syllable_composite}, Appendix), while graphs from conflict contexts revealed more interconnected, complex structures.

\section{Conclusions \& Discussion}

We contributed with an unsupervised pipeline to quantify repertoire and syntax in a graded vocal system, using fruit bats as a case study. Our key finding is that communicative complexity, measured through Maximal Repeats (MRs) and network analysis, is higher in conflict contexts than in cooperative ones.

The finding that temporal compression aids cluster separation aligns with the nature of graded systems, where information is encoded in continuous acoustic modulation. We speculate that basic frequency-based utterances combine and are modulated in time to form more complex syllables, which are then assembled into sequences governed by combinatorial patterns (revealed by MRs) to convey behavioral meaning. We interpret our results through the lens of social complexity. Contexts like Mating Protest and Fighting likely may represent scenarios of social disagreement, requiring more complex signals to negotiate interactions. This is reflected in longer MRs with non-permuted counterparts, and small-world network structures within the syllabic transition graphs. 

We propose the interpretation that higher-complexity observed in conflict-related communication may reflect lower compressibility of information conveying disagreement. We propose to test the use of MRs in other species as a proxy of combinatorial capacity.

\section*{Acknowledgments and Disclosure of Funding}
This paper is adapted from work conducted during the author's Master's thesis.  
The author is grateful to Tim Sainburg for providing valuable feedback 
after the completion of that work. All interpretations and any errors remain the 
author’s own.

The research presented in this paper received no dedicated external funding.
The author gratefully acknowledges financial aid provided from the Earth Species Project, 
and from the Department of Computer and Systems Sciences at Stockholm University 
that support the presentation of this work.

\section*{Code Availability}
The repository associated with the original Master's thesis, on which this paper is based, 
is available at:
\url{https://github.com/gg4u/decodingNonHumanCommunication}

The repository contains the unrefactored thesis implementation. A cleaner and updated version is under development.

\clearpage 
\newpage

\bibliographystyle{plain}
\bibliography{my_neurips_2025}

\newpage
\appendix

\section{Technical Appendices and Supplementary Material}

\newpage

\newpage

\renewcommand{\arraystretch}{1.3}
\begin{table}[htbp]
\centering
\caption{Predictors used in ML classifiers -- Adapted from: \cite{zhang2019comparing}}
\label{tab:rf_features}
\scriptsize
\begin{tabular}{p{0.6cm}p{4.4cm}p{6.6cm}}
\toprule
\textbf{ID} & \textbf{Description} & \textbf{Formula} \\
\midrule
a & Syllable richness & Total number of syllable types in a sequence \\
b & Sequence length & Total number of any syllable-type in a sequence \\
c & Transition count & Total number of bi-gram transitions in a sequence \\
d & Linearity index & $a / c$ \\
e & Contextual variety & Total number of transition types under the behavioral context \\
f & Sequence entropy & $H = -\sum_i p_i \log p_i$ \\
g & Pattern commonness & $\prod_i p(s_i)$ \\
h & Contextual transition strength & $\prod_i p(t_i)$ \\
i & Versatility ratio & Ration between features: $a / b$ \\
j & Transition uncertainty & Entropy of transition probabilities \\
k & Graph transition strength & Product of transitional probabilities (occurring in any context) \\
l & Local predictability & $p_{\mathrm{cond}} \log(p_{\mathrm{cond}})$ \\
m & Global frequency weight & $p_{\mathrm{trans}} \log(p_{\mathrm{trans}})$ \\
n & Conditional syllable chain & $\prod_i p_{\mathrm{cond}}(s_i)$ \\
o & Conditional bi-gram chain & $\prod_i p_{\mathrm{cond}}(B_i)$ \\
p & Transitional bi-gram chain & $\prod_i p_{\mathrm{trans}}(B_i)$ \\
q & 2-step Markov predictability & $\prod_i p(B_i \mid B_{i-1})$ \\
r & Sequence perplexity & $\left(\prod_{i=1}^{N} \frac{1}{p_{\mathrm{cond},i}}\right)^{1/N}$ \\
\bottomrule
\end{tabular}
\end{table}

\newpage

\newpage

\begin{table}[htbp]
\centering
\caption{Parameters used for audio preprocessing.}
\label{tab:audio_preprocessing_params}
\scriptsize 
\begin{tabular}{p{3cm}p{6cm}p{3cm}}
\toprule
\textbf{Function} & \textbf{Description and Best Practices} & \textbf{Settings} \\
\midrule
Bandpass & Cutoff for low and high frequencies & \texttt{low\_freq = 256} \\
         & & \texttt{high\_freq = 120000} \\
\addlinespace[0.2cm]
Noise-Removal & Non-stationary noise removal & \texttt{time\_constant\_s = 0.2} \\
              & & \texttt{time\_mask\_smooth\_ms = 5} \\
              & & \texttt{stationary = False} \\
              & & \texttt{freq\_mask\_smooth\_hz = 256} \\
\addlinespace[0.2cm]
Pre-Emphasis & Emphasis high-frequencies & \texttt{pre\_emphasis = 0.97} \\
\addlinespace[0.2cm]
Short-Time Fourier Transform (STFT) & STFT for power-to-decibel spectrograms. Use a n\_fft of 8ms, with a window of 4ms and 1ms overlaps. Normalize with respect to median power values. & \texttt{n\_fft = 2048} \\
                                    & & \texttt{fmin = 256} \\
                                    & & \texttt{fmax = 120000} \\
                                    & & \texttt{hop\_length = 256} \\
                                    & & \texttt{win\_length = 1024} \\
                                    & & \texttt{sr = 250000} \\
\addlinespace[0.2cm]
Mel-Frequency Cepstral Coefficients (MFCCs) & Use 64 mel-bins & \texttt{as in STFT} \\
                                            & & \texttt{n\_mels = 64} \\
\addlinespace[0.2cm]
Dynamic Threshold Segmentation & Tokenize original audio segments into shorter sub-components & \texttt{as in STFT} \\
                               & & \texttt{db\_delta = 5} \\
                               & & \texttt{ref\_level\_db = 20} \\
                               & & \texttt{pre\_emphasis = 0.97} \\
                               & & \texttt{min\_silence\_for\_spec = 0.1} \\
                               & & \texttt{max\_vocal\_for\_spec = 1} \quad (\# second) \\
                               & & \texttt{min\_level\_db = -60} \quad (\# threshold of sound or noise) \\
                               & & \texttt{silence\_threshold = 0.1} \\
                               & & \texttt{verbose = True} \\
                               & & \texttt{min\_syllable\_length\_s = 0.01} \\
                               & & \texttt{spectral\_range = [2000, 60000]} \\
\addlinespace[0.2cm]
MEL-Filterbank & Compute Log-MEL-Spectrograms. Use a MEL-filter bank to increase the frequency resolution of spectrograms to 4093 (fft\_length // 2 + 1); map them into 32 mel\_bins; increase the relative distance of decibels to 120 db & \texttt{fft\_size = 8192} \quad (\# samples per frame) \\
               & & \texttt{hop\_size = 8192} \quad (\# samples to step) \\
               & & \texttt{fft\_length = 8192 * 2} \quad (\# size of the FFT) \\
               & & \texttt{n\_mels = 32} \\
               & & \texttt{f\_min = 500} \\
               & & \texttt{f\_max = 120000} \\
\bottomrule
\end{tabular}
\end{table}

\newpage

\newpage

\newpage

\begin{table}[h]
\caption{Graph metrics for syllabic transition networks across behavioral contexts for Bat\#215. Metrics include sequence support (number of transitions), small-world coefficients (Sigma, Omega), maximal clique statistics, graph density, and average clustering coefficient (Avg C). A Sigma ($\sigma$) > 1 and Omega ($\omega$) $\approx$ 0 indicates small-world structure.}
\label{tab:graph_metrics_bat215}
\centering
\begin{tabular}{lcccccccc}
\toprule
Context & Support & $\sigma$ & $\omega$ & \# Big Clique & \# All Clique & Density & Avg C \\
\midrule
Biting & 292 & 1.02 & 0.05 & 9 & 99 & 0.40 & 0.46 \\
Feeding & 96 & 0.84 & 0.53 & 4 & 42 & 0.15 & 0.13 \\
Fighting & 50 & 1.00 & 0.03 & 4 & 12 & 0.26 & 0.44 \\
Grooming & 48 & 1.15 & 0.65 & 4 & 35 & 0.11 & 0.09 \\
Isolation & 78 & -- & -- & 2 & 11 & 0.10 & 0.00 \\
Kissing & 48 & 0.86 & 0.63 & 4 & 39 & 0.13 & 0.12 \\
Mating Protest & 629 & 1.00 & 0.00 & 17 & 25 & 0.81 & 0.62 \\
Threat-like & 46 & 1.08 & 0.10 & 5 & 37 & 0.18 & 0.35 \\
\bottomrule
\end{tabular}
\end{table}



\end{document}